\documentclass[conference]{IEEEtran}
\IEEEoverridecommandlockouts
\usepackage{cite}
\usepackage{amsmath,amssymb,amsfonts}
\usepackage{algorithmic}
\usepackage{graphicx}
\usepackage{textcomp}
\usepackage{xcolor}
\usepackage {stfloats} 
\def\BibTeX{{\rm B\kern-.05em{\sc i\kern-.025em b}\kern-.08em
    T\kern-.1667em\lower.7ex\hbox{E}\kern-.125emX}}
\begin{document}
\title{Technical Report of 2023 ABO Fine-grained Semantic Segmentation Competition}
\author{\IEEEauthorblockN{Zeyu Dong}
\IEEEauthorblockA{\textit{University of Glasgow} \\
\textit
2721829d@student.gla.ac.uk}}
\maketitle

\begin{abstract}
In this report, we describe the technical details of our submission to the 2023 ABO Fine-grained Semantic Segmentation Competition, by Team "Zeyu\_Dong" (username:ZeyuDong). The task is to predicate the semantic labels for the convex shape of five categories, which consist of high-quality, standardized 3D models of real products available for purchase online. By using DGCNN as the backbone to classify different structures of five classes, We carried out numerous experiments and found learning rate stochastic gradient descent with warm restarts and setting different rate of factors for various categories contribute most to the performance of the model. The appropriate method helps us rank 3rd place in the Dev phase of the 2023 ICCV 3DVeComm Workshop Challenge.
\end{abstract}

\section{Introduction}
In e-commerce, 3D image semantic segmentation is of great significance. E-commerce platforms can create more vibrant and realistic product presentations by merging 3D photos with semantic segmentation technologies. In order to better grasp a product's design and features, users can rotate, zoom, and browse products in an interactive way, which lessens issues with information asymmetry in online shopping. Additionally, 3D picture semantic segmentation enables customers to edit products and see a real-time preview of the personalising effects in a virtual environment. Different components and features of the product can be highlighted through semantic segmentation. Users can make more informed purchases with the aid of better product information and visualization. Users can lessen returns because products don't live up to expectations, eliminate discontent after purchase, and better comprehend the qualities of the product.\\
This project code link is: https://github.com/ZeUDong/2023-ABO-Fine-grained-Semantic-Segmentation-Competition\\
The competition link  is: https://eval.ai/web/challenges/challenge-page/2027/overview

\section{2023 ABO Fine-grained Semantic Segmentation Competition}
The 3D models used to train and test the model are part of the Amazon Berkeley Objects (ABO) Dataset, which features real objects that can be bought online and are of high quality. These models were expertly designed by artists, and they are made up of build-aware connected components that reflect different form aspects like texture, motion, function, interaction, and construction. The main goal of the workshop challenge is to name the connected components in the ABO dataset with fine-grained semantic labels. As seen in the figure below, the 3D models with build-aware connected components are represented as a collection of convex shapes \cite{yu2023hal3d,collins2022abo}.

\section{Proposed method}
\subsection{DGCNN}
Dynamic Graph Convolutional Neural Network (DGCNN) is a deep learning model for point cloud processing and semantic segmentation, whose main concept is to handle point cloud data using graph convolutional networks (GCN) \cite{manessi2020dynamic,wu2019simplifying}. The adjacency links between the points in the point cloud must be determined in order to construct a graph structure. This can be accomplished by measuring the separation or connectedness between each point and the points close by. Typically, a graph or adjacency matrix is built using this data, with each point being connected to those of its neighbors.
\subsection{SGDR}
We use Stochastic Gradient Descent with Warm Restarts (SGDR) as the Learning rate adjustment strategy, which is a strategy for scheduling cyclic learning rates that is intended to increase the stability and generalizability of model training.\\
Cosine annealing scheduling, which is the main component of SGDR, is used to modify the learning rate. The learning rate fluctuates during the course of the training process, executing periodic annealing in the form of a cosine function. This type of cosine annealing learning rate scheduling starts out with a high learning rate, then steadily drops until it eventually approaches zero. It helps the model converge more quickly in the initial stages of training and then carry out more precise learning in later stages.\\
The learning rate is reset to its initial value at the conclusion of each cosine annealing period, and training is then continued in a new epoch. This occasional restart aids in breaking out of local minima and encourages the model to continue exploring a larger parameter space while being trained. Every cycle is a multiple of the one before it, and they all get longer with time. To better balance the demands of quick convergence and fine-grained model adjustment, this feature enables the learning rate to have variable adjustment speeds at various training phases.
\subsection{Training Pipeline}
We conducted distinct training sessions based on five categories and varying dropout levels for the model training portion. The results for each of the five categories are then optimized separately, increasing the accuracy of the overall results. The learning rate adjustment strategy of all models is SGDR, according to the result of experiments the best dropout for the class chair is 0.6, and the best dropout for others is 0.4.
\begin{figure*}[htbp]
    \centering
    \includegraphics[scale=0.45]{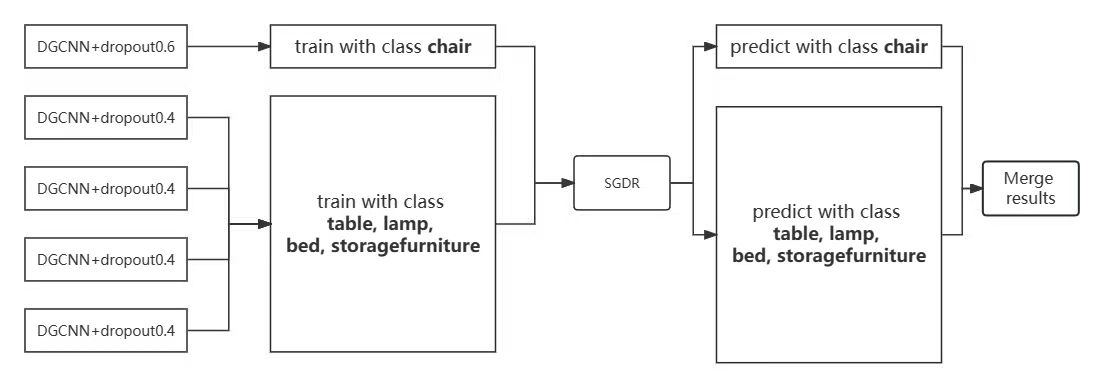}
    \caption{Pipeline}
\end{figure*}
\subsection{Dataset}
The main goal of the workshop challenge is to give connected components in the ABO dataset fine-grained semantic labels. The convex forms used to depict the 3D models with build-aware connected components include. The data is 3D images of different parts of five classes of objects, which contain chair, bed, lamp, storage furniture, and table. All the images are well processed and are split into train, test, and dev. What we need to do is create five distinct models that correlate to various object component classification categories. Then combine the results of these models and output them to the submission file.

\section{Experiment}
\subsection{Evaluiation metrics}
In this competition, the evaluation will be conducted from two aspects, accuracy and Intersection over Union (IoU). Accuracy measures the proportion of samples correctly classified by the model, accuracy = (number of correctly classified samples) / (total number of samples). IoU measures the degree of overlap between the area predicted by the model and the real area, IoU = (intersection area of prediction regions) / (union area of prediction regions).

\subsection{Implementation Details}
According to the baseline offered \cite{yu2023hal3d}, the optimizer is Adam, and the total number of epoch is 250. The learning rate decay is to multiply the factor, which is 0.8, every 25 epoch. The learning rate is 0.001, and the scene per batch train is 2. The loss of baseline is 0.156. \\
At first, we changed the number of epoch to 300 at first, the loss is 0.0621 and the LB is 0.77 to 0.81. Then, we use SGDR to adjust the learning rate, we just use a single cycle, the number of epoch is 250, the loss is 0.019, and the LB is 0.77 to 0.82. We did several experiments based on the consequence of the model using SGDR and changed the number of epoch and dropouts to improve the performance. Finally, the dropout of chair is 0.6 and the dropout of other classes is 0.4.\\
We used a 3090ti graphics card with 24G video memory to train the model. The training time for a single category was 3.5 hours, and the training time for five categories was 17.5 hours. The time to infer a single graph is 3.2 ms.

\section{Conclusion}
In this challenge, Five models were trained to correspond to various categories, and the best five were pooled. On the ABO dataset, we highlighted the significance of our suggested model, and we nearly met SOTA performance. Along with the above-mentioned efficient techniques, we also experimented with a number of novel techniques during the participation process, such as batch size reduction and changing Relu to LeakyRelu. These techniques, however, will not lead to better performance.\\
For future works, we would try to use other powerful backbones, such as a 3D transformer \cite{bas20173d}, to test whether the performance can be improved. We believe that by using the appropriate method as we did in this challenge the accuracy as well as the IoU would be enhanced and improved.

\end{document}